\documentclass[11pt,a4paper]{article}
\usepackage{amsmath}
\usepackage{fancyhdr}
\usepackage{graphicx}
\usepackage{times}

\usepackage{hyperref}
\hypersetup{
    colorlinks=true,
    citecolor=black,
    anchorcolor=black,
    linkcolor=black,
    filecolor=black,      
    urlcolor=black,
    }
\usepackage{tikz}
\usepackage{xcolor}
\definecolor{pyclassblue}{RGB}{85,45,130}
\newcommand{\pyclass}[1]{\textcolor{pyclassblue}{\texttt{#1}}}

\usepackage[left=2.5cm,right=2.5cm,top=3.8cm,bottom=3.0cm]{geometry}

\makeatletter
\renewcommand{\section}
{\@startsection{section}{1}{0mm}{-\baselineskip}{1mm}{\large \bf}}
\renewcommand{\subsection}
{\@startsection{subsection}{1}{0mm}{-\baselineskip}{1mm}{\normalsize \bf}}
\renewcommand{\subsubsection}
{\@startsection{subsubsection}{1}{0mm}{-\baselineskip}{1mm}{\normalsize \it}}
\makeatother

\newlength{\pagenumbershift}
\def\thepage{\textit{\arabic{page}}}

\usepackage{setspace}

\usepackage{titlesec}

\titleformat*{\section}{\large\bfseries\parskip=-5pt}
\titleformat*{\subsection}{\normalsize\bfseries}

\usepackage{caption}
\titlespacing*{\section}{0pt}{15.5pt}{-4pt}
\titlespacing*{\subsection}{0pt}{5pt}{-4pt}
\titlespacing*{\subsubsection}{0pt}{5pt}{-4pt}

\newlength{\bibitemsep}


\fancypagestyle{copyrightfooter}{%
  \fancyhf{}
  
  \fancyfoot[L]{
  \small \em \vspace{-35pt} 
  \rule{\textwidth}{0.35pt}\vspace{-1pt} 

  Copyright \textcopyright\,2026 by 
  Ricardo Maia Avelino, Mennatallah El-Assady, Tom Van Mele, Philippe Block  

  Published by the International Association for Shell and Spatial Structures (IASS) with permission.
  }
}

\begin{document}
\setlength{\baselineskip}{13pt}
\thispagestyle{copyrightfooter}

\begin{flushright}
\ \\
\vspace{-1.9cm}
\footnotesize{\em  Proceedings of IASS-IWSS 2026} -- September 14-18, 2026, Turin, Italy\\
{\em ``R-Evolution of Shapes: Sustainability, Re-Use, and New Design Paradigms''}\\
Annual Symposium of the International Association for Shell and Spatial Structures\\
3$^{\rm rd}$ Italian Workshop on Shell and Spatial Structures\\
\footnotesize{A. Manuello Bertetto, S. Gabriele, F. Marmo, A. Micheletti (eds.)}
 \vskip-10pt
 \rule{\textwidth}{0.4pt}
\end{flushright}

\begin{tikzpicture}[remember picture,overlay]
\node[xshift=23.2mm,yshift=-22.9mm,anchor=north west] at (current page.north west)
{\includegraphics[width=47mm]{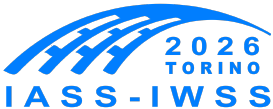}};
\end{tikzpicture}

\begin{center}
\vspace{-15pt}
{\bf \fontsize{16}{19}\selectfont 
Can AI Co-Design Structures? \\An AI Agent for Early-Stage Structural Design
} 

\vspace{0.2cm}

Ricardo MAIA AVELINO*$^{\,\mathrm a, \mathrm b}$, Mennatallah EL-ASSADY$^{\,\mathrm b}$, Tom VAN MELE$^{\,\mathrm a}$, Philippe BLOCK$^{\,\mathrm a}$
\\

\vspace{10pt}

{
\small
$^{*}$Stefano-Franscini-Platz 1, HIB E 45, Zurich 8093, Switzerland\\
mricardo@ethz.ch
}

\vspace{5pt}

{
\small
$^{\,\mathrm a}$
Block Research Group, Department of Architecture, ETH Zurich
\\ 
$^{\,\mathrm b}$
IVIA Lab, Department of Computer Science, ETH Zurich
}

\vspace{-5pt}

\end{center}
\noindent
{\bf \large Abstract}\\[2pt]
Recent advances in large language models and agentic workflows offer a new paradigm to early-stage structural design translating design intent into structural models. Although text-to-3D approaches have demonstrated that natural language can be used to produce visual, mesh-based geometric models, structural design requires domain-specific geometric arrangements and semantically rich data types. This paper introduces a proof-of-concept workflow for prompt-to-structure generation using an AI agent connected to a COMPAS-based structural modelling API through a Model Context Protocol (MCP) server. Through controlled modelling operations, the agent creates and modifies a graph-based structural data model composed of explicit building elements. The resulting artifacts can be inspected, modified, serialised, and further connected to downstream AEC workflows. The approach is demonstrated through selected prompt-to-structure examples and by progressive and editable design sequences that resemble typical iterative structural design workflows. The results are preliminary, but demonstrate the possibility of combining natural language, agentic workflows, and domain-specific structural data structures as a first step toward AI-assisted structural co-design.

\begin{spacing}{1}
\vspace{5pt}\small
{\bf Keywords}: 
Structural Design, Artificial Intelligence, AI agents, Co-design, Human-Computer Interaction

\end{spacing}

\vspace{3pt}

\section{Introduction}

Structural design is not only a process of satisfying performance constraints but also a creative discipline in which form, force, materials, and construction logic are negotiated together. Torroja emphasized the role of intuition and imagination in the conception of structural form \cite{torrojaPhilosophyStructures1958}, while Billington described structural art as the integration of efficiency, economy, and elegance \cite{billington1983tower}. From this perspective, early-stage structural design cannot be reduced to the automatic generation of final solutions, but rather is an iterative process in which designers progressively develop their final result.

Previous work on computational structural design and engineering education has focused on making this exploration more creative and interactive by introducing geometry-based approaches to structural understanding \cite{teng2004fostering, allenFormForcesDesigning2010, vanmele2012geometry, maiaavelinoInteractiveImplementationAlgebraic2021}, developing computational tools to explore families of structural forms \cite{ohlbrock2016cem, lee2016automatic, saldanaochoa2021beyond} and performing topology exploration rather than just looking for a single optimal design \cite{mueller2014computational, OVAL2019185, danhaive2021design}. Further developing this line of work, text-enhanced structural design systems have introduced language as an interface for form-finding and conceptual exploration \cite{guo2022text2form3d}, including graph- and transformer-based algorithms \cite{bleker2026text2structure3d}. In this context, recent developments in large language models (LLMs) and agentic workflows offer a new paradigm to early-stage structural design where AI systems can better understand design intent, create and edit 3D artifacts towards a \textit{co-design} process. 

Recent text-to-3D systems have made substantial progress in generating visual 3D assets represented as neural fields, meshes, or learned 3D latents \cite{poole2022dreamfusion, lin2022magic3d, xiangStructured3DLatents2025}. More recent work in LLM-assisted modelling has begun to explore the generation of 3D CAD geometries and 2D FEA analysis from written scripts \cite{du2024blenderllm, lu2025ll3m, li2025llm4cad, khan2024text2cad, liangIntegratingLargeLanguage2025}. These developments suggest that language can become an increasingly powerful interface for geometry generation. However, structural design requires more than a generic mesh-based structural geometry as they must contain semantically rich objects, including joints, frames and areas, connectivity information, and domain-specific metadata.

Building on the need to develop interactive design processes and on the emerging capabilities and potential of LLMs and agentic systems, this work proposes a workflow that connects a language-model agent to a domain-specific structural modelling API through a Model Context Protocol (MCP) server. The agent interprets natural-language design intent and requests modelling actions that generate or edit a COMPAS-based graph data structure \cite{compas-dev} that contains explicit structural objects such as joints, frames, slabs, walls, grids and levels. This object-based representation allows the generated artifact to be inspected, modified, serialized, and further connected to downstream AEC workflows.

The contribution of this paper is a proof-of-concept workflow for early-stage structural modelling with language agents. It introduces a COMPAS-based structural data structure for representing generated buildings as editable graph-based artifacts, and it demonstrates the approach through selected prompt-generated examples, a language-based editing sequence, and a progressive modelling example. By generating explicit structural data that can be inspected, modified, and connected to downstream workflows, this work paves the way to further research on interactive natural-language-based interfaces for structural design.

\section{An Agentic Structural Modelling Workflow}

\begin{figure}
    \centering
    \includegraphics[width=0.99\textwidth]{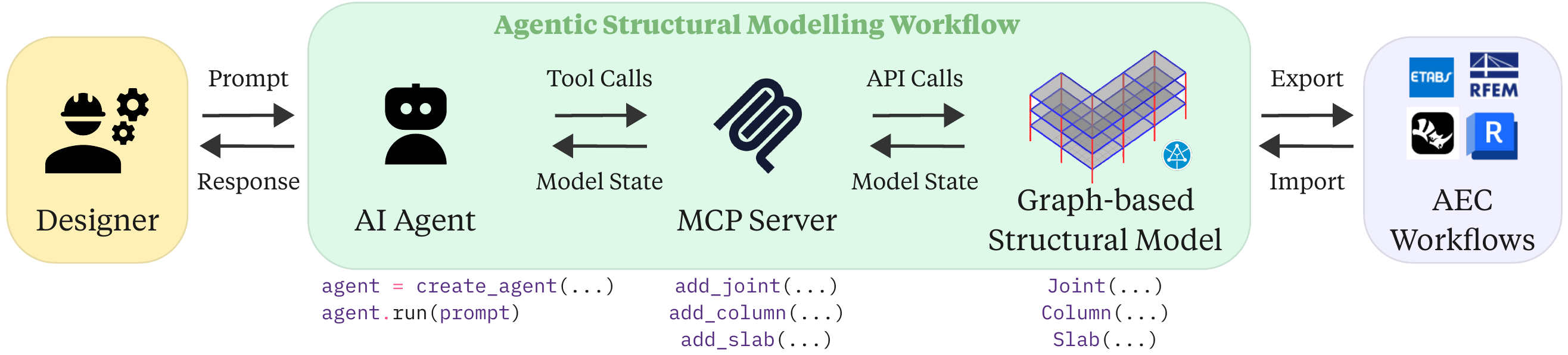}
    \caption{Agentic structural modelling workflow. A designer provides a natural-language prompt to an AI agent, which translates the prompt to MCP tool calls. The calls generate or modify a Graph-based Structural Model. The resulting artifact can be inspected, modified and further exported into AEC workflows}
    \label{fig:workflow}
\end{figure}

Figure~\ref{fig:workflow} summarizes the proposed workflow. A designer provides a natural-language prompt describing a desired structural model or modification. A specialised AI agent interprets this request and communicates with a structural modelling environment through a Model Context Protocol (MCP) server. The server acts as an interface between natural-language reasoning and deterministic modelling actions. These actions generate or modify a 3D structural model. The resulting model can then be reviewed and further adjusted using follow-up instructions. The individual components of this workflow are detailed below.

\subsection{A COMPAS-based graph data structure}

The central component of the workflow is the domain-specific data structure. The generated artifact is represented as a COMPAS \cite{compas-dev} graph-based model which defines the main elements of the system, their connectivity and key meta-data associated with them. These elements are semantically rich and directly relate to the definitions used in structural engineering practice. 


The structural API exposes a small set of Python classes corresponding to common building-scale structural elements and enable operations on them such as additions, deletions, copy, translations, and rotations. The API enables the creation of \pyclass{Joints()} as points in the 3D space, \pyclass{Frame()} connecting two joints, and \pyclass{Area()} elements connecting 3 or more points. The \pyclass{Frame()} object is further specialised into \pyclass{Beam()}, \pyclass{Column()} and \pyclass{Brace()} to represent horizontal, vertical, and inclined elements in the structure. The \pyclass{Area()} object is further specialised in \pyclass{Wall()} and \pyclass{Slab()} elements. Classes for storing metadata assigned to these elements are also present in the API, including \pyclass{Material()} and \pyclass{Section()}. Geometry constructors are also present and allow for the definition of \pyclass{Grid()} and \pyclass{Level()} as well as to operate geometrically on these entities.

These objects are assembled in a \pyclass{Structure()} class, which stores a graph-based representation of the structural model that enables direct access to adjacent entities. In this graph, the nodes correspond to the joints, whereas the frame and area elements encode relationships among those nodes. The API facilitates common operations in building-scale modelling, including creating and modifying regular grids, adding and replicating stories, and copying and dragging objects. This API is exposed to the AI agent through an MCP server, so the operations are executed through function calls as defined below.

\subsection{MCP as a bridge between AI agent and structural model}

Recently, the Model Context Protocol (MCP) was introduced \cite{anthropic2024mcp} as an open protocol for connecting AI assistants to external tools, data sources, and services. In AEC, recent work has begun to use MCP-like interfaces to allow LLMs to query, generate, and modify IFC or BIM models \cite{nithyanantham2025mcp4ifc, heimigelschner2026mcpbim} and generate OpenSees and ETABS structural analysis models \cite{avilaHumanAITeaming2025}. 

In the present work, an MCP server is created to expose the COMPAS-based structural modelling API to the agent. When triggered by the AI agent, the MCP functions execute deterministic operations and return the updated state of the model. These controlled actions populate and modify a graph-based structural data structure. As a result, an explicit 3D structural artifact is produced that can then be inspected visually and modified through follow-up instructions. The AI agent is pre-configured to use the MCP definitions to build the structural artifacts and relies on Google Gemini 3.1 Pro \cite{Gemini31Pro} for reasoning on the user input. 

\subsection{Import and Export}

The generated structures are stored as explicit COMPAS objects whose data can be serialised and stored as a structured JSON file. Based on the COMPAS ecosystem interoperability features, the geometric data underlying the model can be imported and exported to common geometry formats. The development of different connectors to common AEC software is planned as future work such that the generated models could be directly transformed into AEC software models typically used in structural practice.

The next section demonstrates the use of the presented workflow to generate, edit, and progressively design complex structural building artifacts that can serve engineers and architects working on early-stage structural design.

\section{Prompt-to-Structure Demonstrations}

\subsection{Generating 3D Structural Models from Natural Language}

\begin{figure}[h!]
    \centering
    \includegraphics[width=0.9\textwidth]{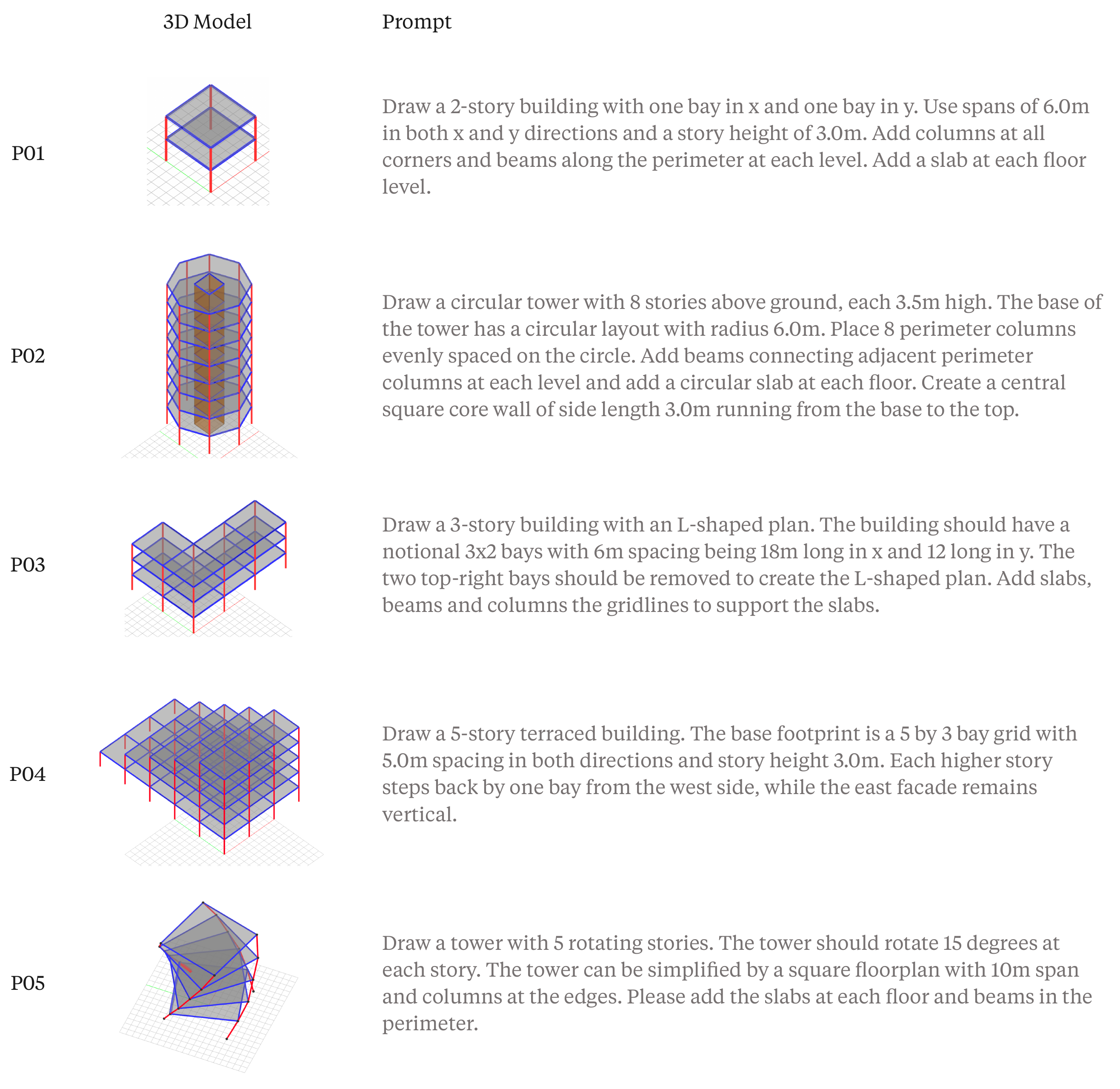}
    \caption{Selected prompt-to-structure examples. P01 shows a simple two-story one-bay frame; P02 introduces a circular tower with a central core; P03 generates an L-shaped footprint; P04 introduces a multi-story building with setbacks; and P05 shows a rotating tower. Each model is generated from a direct natural-language prompt and stored as an editable structural artifact.}
    \label{fig:examples}
\end{figure}

Figure~\ref{fig:examples} presents five selected examples of prompt-generated structural models that show the potential of the current workflow to generate a range of structural building models. The examples move from simple to more complex artifacts. P01 demonstrates the basic creation of columns, beams, and slabs from span and height information. P02 introduces a circular plan, repeated perimeter columns, and a core wall. P03 uses a notional grid and removes bays to form an L-shaped plan. P04 extends the task to a terraced multi-story structure with different floor extents across levels. P05 introduces geometric rotating stories along the height of the building. Together, these examples suggest that the agent can generate models that mix different structural arrangements and geometric operations. Furthermore, the agent can handle domain-specific vocabulary used to describe building-scale models.

\subsection{Editing object-based structural models}

\begin{figure}[h]
    \centering
    \includegraphics[width=0.9\textwidth]{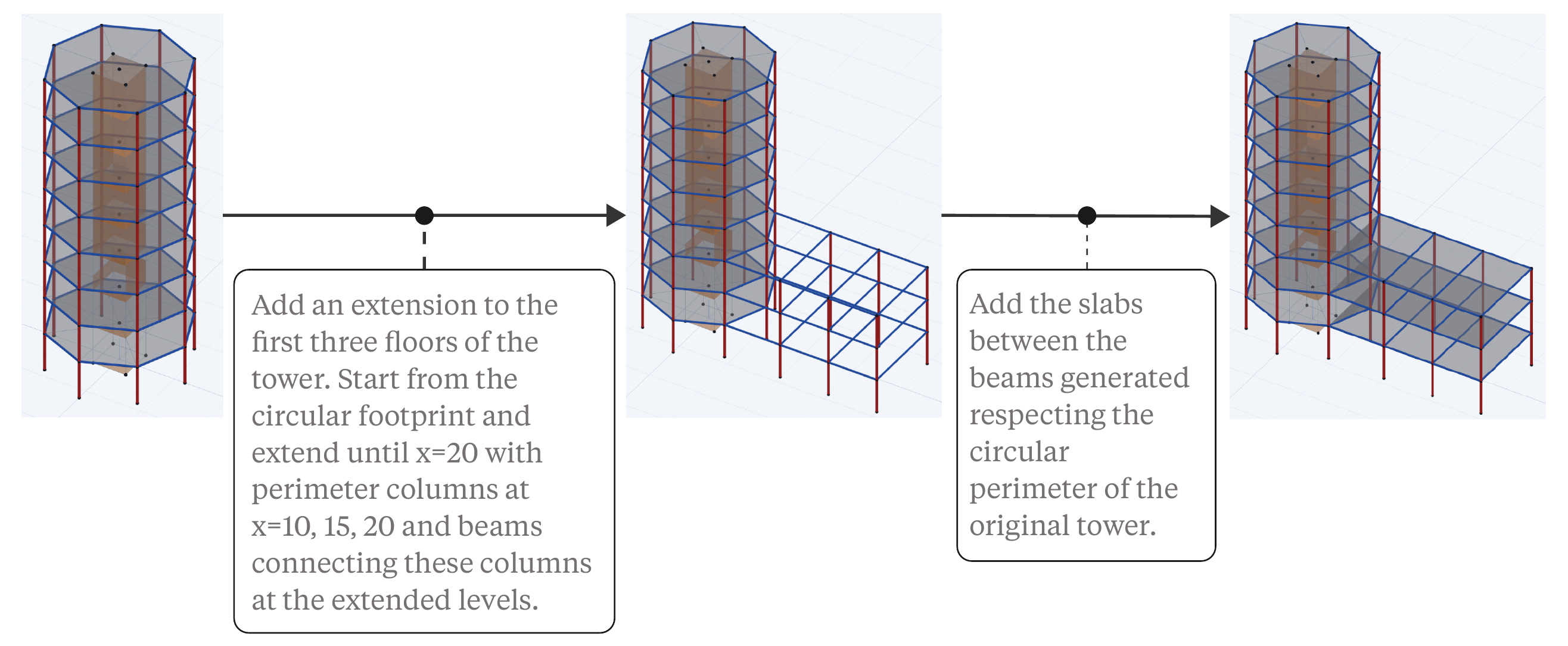}
    \caption{The circular tower generated by prompt P02 has the lower floors extended through a follow-up natural-language instruction.}
    \label{fig:editing}
\end{figure}

Figure~\ref{fig:editing} illustrates how a generated artifact can be modified through follow-up language instructions. Starting from the circular tower in P02, the user asks for an extension to the first three floors. The AI agent first queries about the current state of the model, reasons about the required addition or modification, and the maps the MCP tools that need to be called to achieve it. For this example, the agent responds to the user prompt by adding beams and columns creating the extension as requested. The user then follows up with another modification, requesting the addition of slabs on the newly created floors. The updated version adds the slab elements to the designated floors. 

This example reflects a common situation in early-stage design, where new ideas or design modifications are proposed to an already existing structure, which then need to be reflected in an updated model. By allowing for this interaction, the workflow caters to the practical needs of structural designers as they adapt their models towards the final design solution.

The example demonstrates that object-level modifications can be requested and executed on an existing model, but also reveals limitations in geometric interpretation. In Figure~\ref{fig:editing}, the agent was unable to correctly draw the slab adjacent to the initial circular footprint as requested, resulting in an unnecessary overlap between the added rectangular slab and the existing central tower slabs. Further prompt instructions or manual edits would then be needed to update this particular location as intended. Geometric errors can indeed occur due to current LLMs' limited spatial reasoning or to ambiguity on the user prompting. This limitation highlights the need for manual inspection and verification.

\subsection{Towards progressive structural co-design}

\begin{figure}[h]
    \centering
    \includegraphics[width=0.9\textwidth]{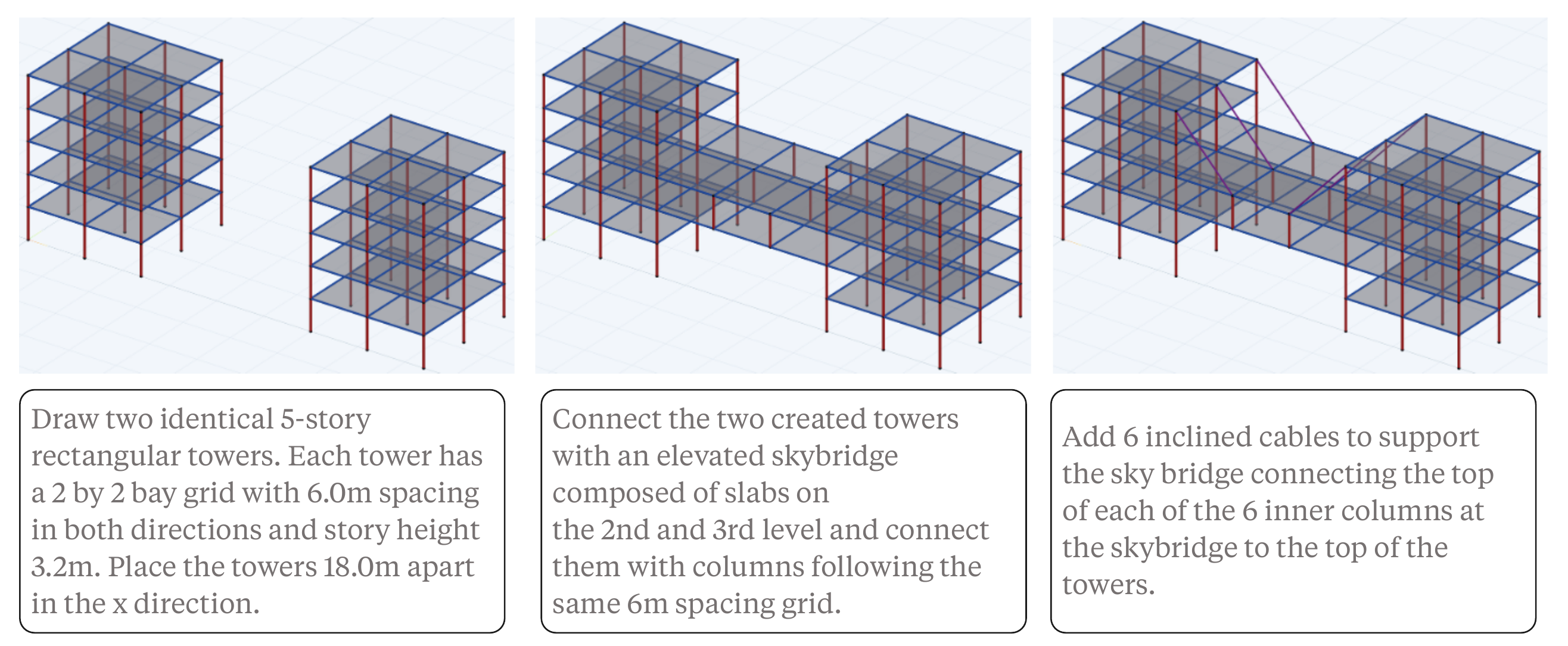}
    \caption{Progressive language-based structural modelling. A designer progressively complexifies the building through a sequence of prompts: two towers are generated, then connected by a skybridge which is later supported with inclined cable elements.}
    \label{fig:cocreation}
\end{figure}

Figure~\ref{fig:cocreation} presents a second sequence in which the designer deliberately starts with a simple artifact and increases its complexity over several prompts. The initial request creates two towers. A subsequent prompt connects the towers through an elevated skybridge, and a later prompt adds inclined cables to support the bridge. This sequence illustrates a different mode of use in which the user does not provide the complete design brief at once, but develops the artifact progressively. Such progressive modelling resembles early structural design practice, where designers often begin with a simple organizational idea and even generate simplified preliminary models that then become more complex over time.

By exposing the AI agent to the model state at all times, this progressive design is possible. This example connects the current workflow to the broader idea of AI-assisted structural co-design where the AI agent is able to rapidly materialize intermediate structural artifacts, allowing the designer to inspect, revise, and redirect the model. This generation and edition of the models is a prerequisite for future structural co-design workflows.

By reducing the effort required to create preliminary structural models and removing the friction of structural modelling, such workflows can support teams to achieve higher-performing design alternatives. The opportunities and limitations of the findings presented here are discussed in the next section.

\section{Opportunities and Limitations}

The workflow presented suggests several opportunities for early-stage structural design. First, prompt-based model generation can reduce the friction associated with repetitive model setup, allowing designers to test structural ideas more quickly. By lowering the effort required to create preliminary models, the workflow can enable greater design freedom and unlock more exploratory forms of creativity. Second, the use of explicit structural elements and access to the current state of the model using MCP servers make such that the resulting artifact can be inspected, modified and exported. Third, future AI-assisted modelling could support the design of more resource-efficient or sustainable structures by guiding the user on modifications, additions, or deletions that can have a positive outcome on the structure metrics, including material consumption and embodied carbon accounts.

The approach may also be extended through multimodal input. Structural designers and architects do not communicate through text alone: sketches, annotations, reference images, or preliminary models are also important \cite{Avelino2026_cocreation}. Future systems could therefore combine these multimodal inputs to bring human and AI understanding closer during model interaction.

The current prototype also has important limitations. The examples shown in this paper are proof-of-concept rather than a formal evaluation. The generated models do not guaranty structural correctness, code compliance, or analysis readiness. Prompt ambiguity, hallucination, misinterpretation of design intent and robustness across similar prompts still needs to be assessed. Furthermore, current LLMs are known to have limited spatial reasoning \cite{yangThinkingSpaceHow2025}, such that geometric modifications can be wrongly executed by agents. The size and complexity of the models studied here are also limited. These limitations motivate the need for further research and development of benchmarks, datasets, and validation protocols for AI-assisted structural modelling. 

\section{Conclusions}

Building on existing research that formalizes structural design as an exploratory and creative discipline, this paper presented a proof-of-concept workflow to generate editable early-stage structural building models from natural-language prompts. The workflow presented combines language-agent reasoning with a COMPAS-based structural modelling API exposed through an MCP server. Through selected examples, the paper shows that language agents can create editable structural artifacts that can assist in early structural design workflows. This work argues that LLMs introduce a new design paradigm in structural design and make it possible to establish new modes of interaction between human designers and structural models. The contribution is preliminary and a series of limitations exist regarding the size of the models, analysis readiness, and the risk of intent misinterpretation. Nevertheless, the results suggest that structural design is a relevant and challenging domain for future AI-assisted modelling workflows, where the goal is not autonomous design but the reduction of friction between design intent and editable structural models towards more creative and efficient designs.

\section*{Acknowledgements}

This research was supported by the Design++ Postdoctoral Fellowship awarded to the first author in collaboration with Halter AG.

\bibliographystyle{iass-doi-url}
\bibliography{bibtex}

\end{document}